\documentclass[11pt,a4paper]{article}
\usepackage[hyperref]{acl2019}

\usepackage{times}
\usepackage{latexsym}

\usepackage{amsmath, amsthm, amssymb, amsfonts}
\usepackage{algorithm}
\usepackage{algpseudocode}

\usepackage{helvet}  %Required
\usepackage{courier}  %Required
\usepackage{url}  %Required
\usepackage{graphicx}  %Required
\usepackage{color}

\usepackage{multirow}
\usepackage{booktabs}
\usepackage{comment}
\usepackage{caption}

\usepackage{natbib}
\usepackage{pinyin}
\usepackage{subfigure}
\usepackage{diagbox}
\usepackage{verbatim}
\usepackage{stmaryrd}
\usepackage{hyperref}
\usepackage{scrextend}
\usepackage[normalem]{ulem}

\newcommand{\tl}[1]{\multicolumn{1}{l}{#1}} %左对齐 

\aclfinalcopy % Uncomment this line for the final submission
\title{Efficient Bidirectional Neural Machine Translation}

\author{Xu Tan$^\S$, Yingce Xia$^\S$, Lijun Wu$^\dag$ and Tao Qin$^\S$ \\
$^\S$Microsoft Research \\
$^\dag$Sun Yat-sen University \\
\{xuta,taoqin\}@microsoft.com  \\
}

\date{}

\begin{document}
\maketitle
\begin{abstract}
The encoder-decoder based neural machine translation usually generates a target sequence token by token from left to right. Due to error propagation, the tokens in the right side of the generated sequence are usually of poorer quality than those in the left side. In this paper, we propose an efficient method to generate a sequence in both left-to-right and right-to-left manners using a single encoder and decoder, combining the advantages of both generation directions. Experiments on three translation tasks show that our method achieves significant improvements over conventional unidirectional approach. Compared with ensemble methods that train and combine two models with different generation directions, our method saves 50\% model parameters and about 40\% training time, and also improve inference speed. 

%Neural machine translation typically leverages the encoder-decoder framework, where the decoder usually generates a target sequence token by token from left to right. Due to error propagation, the tokens in the right side of the generated sequence are usually of poorer quality than those in the left side. In this paper, we propose a novel and efficient framework to generate a sequence with both left-to-right and right-to-left manners in a single encoder and decoder, combining the advantages of both generation directions. Experiments on three translation tasks demonstrate that our proposed method can achieve significant improvements over conventional unidirectional approach. Compared with the conventional methods that train two models with different generation directions for ensembling, our method can save 50\% model parameters, nearly 40\% training time and also improve the inference speed. 
\end{abstract}

\section{Introduction}
\label{sec_intro}
Neural Machine Translation (NMT) is very challenging and attracts lots of research attention in recent years~\citep{DBLP:conf/emnlp/ChoMGBBSB14,DBLP:journals/corr/BahdanauCB14,DBLP:conf/emnlp/LuongPM15,gehring2017convolutional,vaswani2017attention,shen2018dense,he2018layer,hassan2018achieving,song2019mass}. The major approach to the task typically leverages an auto-regressive encoder-decoder framework: The encoder transforms the source sequence into hidden representations, and the decoder generates a target sequence token by token from left to right conditioned on the source hidden representations. 

Once a preceding target token is mistakenly generated, the quality of the current token could be affected, and the error would be further propagated to negatively influence the prediction of future tokens. Therefore, the left-to-right decoder could result in poorly generated tokens in the right side of the sequence. We conduct a statistical analysis on the translation results of a well-trained Transformer~\citep{DBLP:conf/nips/VaswaniSPUJGKP17} model on WMT17 Chinese-English validation set by comparing the prediction accuracy of the first 4 and last 4 tokens of the generated sequence. The accuracy for the first 4 tokens is 39.0\%, while for the last 4 is only about 28.1\%. The same phenomenon was also observed in~\citet{DBLP:conf/aaai/LiuFUS16,wu2018beyond,ren2019almost}. This shortcoming limits the quality of neural machine translation. 

\begin{table*}[t!]
\small
\centering
\begin{tabular}{l c}
\toprule
\multirow{2}{*}{\textbf{SRC}} & \tl{\footnotesize \guo{2}\nei{4} \yi{1}\jia{1} \quan{4}\shang{1} \fen{1}\xi{1}\shi{1} \biao{3}\shi{4}, \dong{1}\fang{1} \zheng{4}\quan{4} \zai{4} \gang{3} \ji{1}\chu{3} \bing{4}\bu{4}\hao{3}, \xiang{1}\bi{3} \zhong{1}\yin{2}\guo{2}\ji{4}, }\\
& \tl{\zhong{1}\xin{4}\zheng{4}\quan{4} \deng{3} \yin{2}\hang{2} \xi{4} \quan{4}\shang{1}, \xian{3}\de{2} \xian{1}\tian{1} \bu{4}\zu{2}.} \\
\midrule
\multirow{2}{*}{\textbf{REF}} & \tl{\footnotesize A domestic brokerage analyst said that the Orient Securities in Hong Kong did not enjoy a good foundation. It has} \\ 
& \tl{\footnotesize many inherent problems compared with Bank of China International and CITIC Securities, etc.} \\
\midrule
\midrule
\multirow{2}{*}{\textbf{L2R}}  & \tl{\footnotesize \uwave{A domestic brokerage analyst said that Oriental Securities does not have a good foundation in Hong Kong}, its} \\
& \tl{\footnotesize capital management is the director of EIC.} \\
\midrule
\multirow{2}{*}{\textbf{R2L}} & \tl{\footnotesize (Citic Securities) A domestic brokerage analyst said that Orient Securities based in Hong Kong is not good , \uwave{which}} \\
& \tl{\footnotesize \uwave{is insufficient compared to bank brokerage houses such as Bank of China International and Citic Securities.}} \\
\bottomrule
\end{tabular}
\caption{\footnotesize Translation examples on WMT17 Chinese-English dataset in L2R (left-to-right) and R2L (right-to-left) generation directions. SRC/REF represents the source/reference sentence. Text highlighted in wave line represents the correct translation.}
\label{tab:case_example}
\end{table*}

A straightforward idea to tackle the aforementioned problem is to consider a right-to-left decoder, which generates the target sequence token by token from right to left and makes more accurate predictions for the tokens in the right side. Table \ref{tab:case_example} shows the translation results of a left-to-right decoder and a right-to-left decoder based on Transformer~\citep{DBLP:conf/nips/VaswaniSPUJGKP17} for a Chinese sentence from the WMT17 Chinese-English translation dataset. It can be seen that while the left-to-right decoder does well on the left side of the target sequence, the right-to-left decoder does well on the right side of the target sequence. Therefore, several works have studied how to combine the left-to-right decoder together with the right-to-left decoder~\citep{DBLP:conf/aaai/LiuFUS16,DBLP:conf/naacl/LiuUFS16,DBLP:conf/wmt/SennrichHB16,DBLP:conf/wmt/WangCJYCLSWY17}. Most of them train two separate encoder-decoder models (i.e., one with a left-to-right decoder and the other with a right-to-left decoder), use one model to generate several candidate sequences, and then leverage the other model to re-rank those candidates. While they can indeed improve the accuracy of sequence generation, they need to pay the price of training an additional model.

In this paper, we propose a simple yet efficient model that just uses a single model (with an encoder and a decoder) to bidirectionally generate the sequence. We optimize the model in both left-to-right and right-to-left generation manner under a unified log-likelihood loss. In order to inform the decoder a sense of generation direction, we learn the embeddings of two start tokens for the decoder, each representing a decoding direction. For example, when using the right-to-left start token as the start input, we reverse the target sequence in the training data, guiding the model to generate the sequence in a right-to-left manner. 

%During inference, we using beam search to generate sequences in both directions and select the sentence that maximizes the unified log likelihood.

Our method, to the best of our knowledge, is the first work to explore bidirectional decoding in a single model (only an encoder and a decoder). Through experiments on three machine translation tasks, we demonstrate that our proposed method can achieve significant improvements over the conventional unidirectional model. In particular, our method achieves a 29.30 BLEU score (see Table~\ref{nmt_ende_stoa} in Section~\ref{sec_nmt_ret}) on WMT14 English-German translation task. In terms of efficiency, our method can save 50\% model parameters, nearly 40\% training time, and also improve the inference speed (see Section~\ref{sec4_analysis}).
We will release the code once the paper is open to the public.

%Compared with the previous works, our method has the following advantages: 1) It is much more light-weight in computation and storage than the separately trained two models; 2) Compared with the works that train left-to-right and right-to-left models with separate losses, the inference in our method is more consistent with the training, since they are guided by the same unified log-likelihood loss, and such a unified loss keeps us free from tuning the hyperparameters in the previous literature for the tradeoff between two losses.

\section{Bidirectional NMT}
\label{sec_3}
In this section, we introduce the general approach of bidirectional NMT, as well as the bidirectional decoding during training and inference. 

\subsection{Method}

To enable bidirectional sequence generation in a single model, we decompose the probability of the target sequence into the products of per-token probabilities in both the left-to-right manner and the right-to-left manner:
\begin{equation}
\small
\begin{aligned}
\label{eq_loss}
&L(S; \theta) = \sum_{({x},{y}) \in S} (\log P(\overrightarrow{y}|x;\theta) + \log P(\overleftarrow{y}|x;\theta)) \\
&= \sum_{({x},{y}) \in S} \sum_{t} (\log P(y_t|{y_{<t}}, x;\theta) + \log P(y_t|{y_{>t}}, x;\theta)),  
\end{aligned}
\end{equation}
where $S$ denotes the training corpus, $P(\overrightarrow{y}|x;\theta)$ denotes the sequence probability generated in the left-to-right manner and $P(\overleftarrow{y}|x;\theta)$ denotes the right-to-left manner. Note that each generation direction is based on the same set of parameters $\theta$. $P(y_t|{y_{<t}}, x;\theta)$ denotes the generation probability of token $y_t$ at position $t$ conditioned on the input $x$ and its left-side tokens $y_{<t}$, and $P(y_t|{y_{>t}}, x;\theta)$ denotes the generation probability of the token conditioned on $x$ and its right-side tokens $y_{>t}$.

Through maximization of the above log likelihood, a token $y_t$ can be generated based on its left-side context and can also be predicted based on its right-side context, and most importantly, using the same parameters $\theta$. That is, our learned model can support both left-to-right decoding and right-to-left decoding.

%In the training phase, we simultaneously maximize the probability of $y_t$ given the context information in the left part and in the right part. It's more beneficial compared with the conventional unidirectional model that only considers one side information in one model.
%, which leads to poor quality on the other side of sequence generation.

The encoder in our proposed method is the same as the encoder $f_{enc}$ in the conventional encoder-decoder framework, which maps the source input into the hidden representation. For the decoder, unlike conventional models using a zero vector as the start token for decoding, we learn two embedding vectors for two start tokens representing the decoding directions, one for left-to-right and the other for right-to-left decoding.

\begin{comment} 
\begin{equation} 
\label{eq_bidirectp}
\small
\begin{aligned}
    P({y_t}|{y_{<t}},{x};\theta) = f_{dec}([\overrightarrow{SOS}, {y_{<t}}],h), \\
    P({y_t}|{y_{>t}},{x};\theta) = f_{dec}([\overleftarrow{SOS}, {y_{>t}}],h), \\
\end{aligned}
\end{equation}
where $\overrightarrow{SOS}$ and $\overleftarrow{SOS}$ are the two learnt start tokens that determine the left-to-right and right-to-left decoding direction, respectively. $h$ is the hidden representation extracted by the encoder and $[\cdot ,\cdot]$ means concatenation.

If the input $x$ is also a sequence of tokens, an alternative and straightforward way to indicate different decoding directions is to add two indicator tokens into the source sequence, e.g., adding at the beginning or the end of the sequence. However, such a way is costly as we need to run encoder twice to extract the hidden representation for different decoding directions. In contrast, our method only needs to encode the source input once, since we add the indicator tokens to target sequence while decoding.
\end{comment}

\subsection{Training and Inference}
In training, for each original data pair $(x, \overrightarrow{y})$, we get an additional right-to-left pair $(x,\overleftarrow{y})$ by reversing the tokens in the target sequence. In each mini-batch, we mix the original training pairs and the right-to-left pairs together. Note that we add different start tokens into the original pairs and the reversed right-to-left pairs to indicate the decoding directions, as discussed in the last subsection.

%\subsection{Bidirectional Generation during Inference}
During inference, given an input $x$, to keep consistent with the training loss function in Equation (\ref{eq_loss}), ideally we need to find the target sequence that maximizes the summation of the token probability conditioned on the left-side and the right-side tokens:
\begin{equation}
\label{eq_inference}
\small
\begin{aligned}
y^{*} = & \arg\max_{y \in \mathcal{Y}} \sum_{t} (\log P(y_t|{y_{<t}}, x;\theta) + \log P(y_t|{y_{>t}}, x;\theta)).
\end{aligned}
\end{equation}
However, we cannot enumerate all possible target sequences, as there are exponentially many candidates. Instead of searching for the exact $y^*$, we adopt an approximate approach and try to find a reasonably good $y$ based on beam search, as used in the inference process of the standard uni-directional decoding. The inference procedure is described in Algorithm~\ref{alg}.

\begin{algorithm}
\small
\caption{\small Inference Procedure}\label{alg}
\begin{algorithmic}[1]
\State \textbf{Input}: Source input $x$, model $\theta$.
\State Generate $2K$ sequences with beam search, each with $K$ sequences in left-to-right and right-to-left decoding, denoted as $\mathcal{BS}(\overrightarrow{y})$ and $\mathcal{BS}(\overleftarrow{y})$.

\State     For each sequence $y$ in $\mathcal{BS}(\overrightarrow{y})$, record the generation probability $P(\overrightarrow{y}|x;\theta)$ and feed the sequence with right-to-left manner to calculate probability $P(\overleftarrow{y}|x;\theta)$.

\State Apply Step 3 to the sequence in $\mathcal{BS}(\overleftarrow{y})$.
\State Choose the best candidate $y^{*}$ among the $2K$ sequences according to Eqn. (2). 
\State \textbf{Output}: $y^{*}$.
\end{algorithmic}
\end{algorithm}

\begin{comment}
As shown in Algorithm~\ref{alg}, we first use our learned model to generate candidate sequences in both directions in Step 2. When performing decoding in different directions, we just need to feed different start tokens at the beginning of the target sequence to control the corresponding generation directions. Then we need the probabilities of a generated sentence in both directions for Equation (\ref{eq_inference}). The left-to-right probabilities for left-to-right decoding as well as the right-to-left probabilities for right-to-left decoding have already been obtained and recorded, and we only need to calculate the probabilities in the other direction, as shown in Step 3 and 4. Step 5 generates the final output sentence. 
\end{comment}

\section{Experimental Results}
\label{exp_nmt}
We conduct experiments on three translation tasks (IWSLT14 German$\to$English, WMT14 English$\to$German and WMT17 Chinese$\to$English) to very the effectiveness of our proposed method. More details about the dataset description and  experimental setup can be found in the supplementary materials (Section 1 and 2).
\subsection{Results}
\label{sec_nmt_ret}

Table \ref{nmt_ende} shows the BLEU scores of our method on three translation tasks compared with the unidirectional baseline, the Transformer. Note that the baseline and our method are of the same NMT model architecture with the same number of parameters. The only difference between them is the training and inference procedure. From this table, we can observe that our bidirectional generation method outperforms the unidirectional baseline by a large margin, with 0.8 and 1.32 point BLEU score improvement on the IWSLT14 De$\to$En and WMT17 Zh$\to$En tasks, and 0.9 points BLEU score improvement on WMT14 En$\to$De task.
%column "L2R" and "R2L" represent our proposed method, but just use left-to-right or right-to-left manner to decode sequence, 
%when just performing unidirectional ("L2R" in the table) sequence generation without considering the loss in the reverse side, our proposed method can match the BLEU scores of conventional left-to-right baseline. "L2R" consistently outperforms the "R2L" results. 

\begin{table}[tb]
\centering
\small
\begin{tabular}{l c  c}
\toprule
 \textbf{Task} & \textbf{Baseline}  &  \textbf{Our method}  \\
%\cmidrule{3-5} 
\midrule
WMT14 En$\to$De & $28.40^{*}$ &  \textbf{29.30} \\ 
%\hline
WMT17 Zh$\to$En & 24.29  & \textbf{25.32} \\
%\hline
IWSLT14 De$\to$En & 32.17 & \textbf{33.15} \\
\bottomrule
\end{tabular}
\caption{Comparison of our method with the unidirectional baseline. For $28.40^{*}$, we reimplemented the baseline for WMT14 En$\to$De, and got 28.36 BLEU score. Therefore, we simply list the score 28.40 reported in the original paper~\citep{DBLP:conf/nips/VaswaniSPUJGKP17}.}
\label{nmt_ende}
\end{table}

% \multirow{2}{*}{\textbf{Task}}&\multirow{2}{*}{\textbf{Baseline}}  & \multicolumn{3}{c}{\textbf{Our Method}}  \\
%\cmidrule{3-5} 
%& &\textbf{L2R} & \textbf{R2L} & \textbf{BSG}  \\
% 28.25 & 27.08
% 24.31 & 23.81
% 32.14 & 31.88
%\hline
%Zh$\to$En(34M) & 25.90 & 25.86 & 25.08 & \textbf{26.48} & 0.58 \\   

\begin{table}[t]
\centering
\small
\begin{tabular}{l c c}
\toprule
\textbf{Task} & \tl{\textbf{Method}} & \textbf{BLEU} \\
\midrule
\multirow{7}{*}{En$\to$De}  & \tl{ByteNet \tiny\citep{DBLP:journals/corr/KalchbrennerESO16}} & 23.75 \\ 
                            & \tl{GNMT+RL \tiny\citep{DBLP:journals/corr/WuSCLNMKCGMKSJL16}} & 24.60 \\
                            & \tl{ConvS2S \tiny\citep{DBLP:conf/icml/GehringAGYD17}} & 25.16 \\ 
                            & \tl{MoE \tiny\citep{DBLP:journals/corr/ShazeerMMDLHD17}} & 26.03 \\
                            & \tl{Transformer (base) \tiny\citep{DBLP:conf/nips/VaswaniSPUJGKP17}} & 27.30 \\ 
                            & \tl{Transformer (big) \tiny\citep{DBLP:conf/nips/VaswaniSPUJGKP17}} & 28.40 \\ 
         \cmidrule{2-3}
                            
                            & \tl{Our method} & \textbf{29.30} \\
\midrule

%\multirow{5}{*}{Zh$\to$En} & \tl{CASICT-DCU\_NMT \tiny\cite{DBLP:conf/wmt/ZhangPHZL17}} & 22.30\textsuperscript{*}  \\
\multirow{4}{*}{Zh$\to$En} & \tl{uedin-nmt  \tiny\citep{DBLP:conf/wmt/SennrichBCGHHBW17}} & 22.90\textsuperscript{*} \\
  & \tl{xmunmt   \tiny\citep{DBLP:conf/wmt/TanWHCS17}} & 23.40\textsuperscript{*}  \\
 & \tl{SogouKnowing-nmt \tiny\citep{DBLP:conf/wmt/WangCJYCLSWY17}}  & 24.00\textsuperscript{*}  \\ 

\cmidrule{2-3}
                            & \tl{Our method} & \textbf{25.32} \\
\midrule
\multirow{5}{*}{De$\to$En} 
                         & \tl{MIXER \tiny\citep{DBLP:journals/corr/RanzatoCAZ15}} &  21.83  \\
& \tl{AC+LL \tiny\citep{bahdanau2016actor}} &  28.53\\
& \tl{NPMT \tiny\citep{huang2017neural}}  & 28.96\\
& \tl{NPMT+LM \tiny\citep{huang2017neural}} & 29.16 \\
& \tl{Dual Transfer Learning \tiny\citep{Wang2018Dual}} & 32.35 \\

\cmidrule{2-3}
                           & \tl{Our method} & \textbf{33.15} \\
\bottomrule
\end{tabular}
\caption{Comparison with previous works. All the BLEU scores marked with * can be found in http://matrix.statmt.org/matrix/systems\_list/1878.}
\label{nmt_ende_stoa}
\end{table}

Table \ref{nmt_ende_stoa} shows the results reported in previous works. On WMT14 En$\to$De task, our model achieves 29.30 in terms of BLEU score, beating the strong baseline of the Transformer model and setting a new record for this dataset. On WMT17 Zh$\to$En dataset, our model outperforms the winning system of WMT17 Zh$\to$En challenge\footnote{The highest score on WMT17 Zh$\to$En challenge (http://matrix.statmt.org/matrix/systems\_list/1878) is 26.40, based on the ensemble of multiple models. We choose the best single model for comparison.} by 1.32 point of BLEU score. On IWSLT14 De$\to$En dataset, our method also achieves 33.15 BLEU score, higher than the previous works. 

\subsection{Analysis}
\label{sec4_analysis}

\paragraph{Comparison on Training and Inference Time}
We analyze the training time of our method compared with the conventional approach. On IWSLT14 De$\to$En task, with one NVIDIA Tesla M40 GPU and a batch size of 4096 tokens, the training speed is 26828 tokens/second. For the conventional approach that training left-to-right and right-to-left in two different encoders and decoders, it needs 2 separate GPUs to reach the same speed. However, our method needs 1.2x iterations to converge. Totally, our method only requires $0.5*1.2=0.6$x computation in training compared with the conventional approach. The results are similar on other tasks such as WMT14 En$\to$De and WMT17 Zh$\to$En.

During inference, the model needs additionally to calculate the probability of the generated sentence in another direction as shown in Algorithm~\ref{alg}. However, the time for calculating the probability is only nearly 10\% of the generation time, which results in a 10\% additional cost compared with unidirectional baseline. Compared with the conventional method with two models in different generation directions, our method just needs to run the encoder once for bidirectionally generating target sentence, which will also save time during inference.

\paragraph{Analyzing of the Contribution of Right-To-Left Decoding}
Someone may argue that the improvements of our model come from the effects similar to self-ensemble~\citep{laine2016temporal,liu2017towards}, but not the bidirectionally generated sequence. According to our statistics on WMT14 En$\to$De test set, for the final generation results, 66.2\% sentences are from left-to-right decoding while 33.8\% are from right-to-left decoding. If we only use sentences generated in left-to-right as candidates and keep other factors unchanged, the final BLEU score drops from 29.30 to 28.73 (the unidirectional baseline is 28.40). This result demonstrates that our improvements are more due to sentence candidates generated in both directions, but not the effects like self-ensemble.

\paragraph{Accuracy of Our Method with Unidirectional Decoding Only}
We analyze how our model performs when decoding with only one direction, and compare our results with unidirectional baseline on the three datasets. Note that we do not use Algorithm~\ref{alg} for inference but just decode with beam search in left-to-right. The comparison results are shown in Table~\ref{tab_unidirect_compare}. It can be seen that our method with only left-to-right decoding nearly matches the accuracy with left-to-right baseline model and there's no obvious accuracy degradation.
\begin{table}[htbp]
\centering
\small
\begin{tabular}{l c c}
\toprule

&\textbf{Baseline} & \textbf{Our method (left-to-right)}\\
\midrule
En$\to$De & 28.40 & 28.25 \\
Zh$\to$En & 24.29 & 24.31 \\
De$\to$En & 32.17 & 32.14 \\
\bottomrule
\end{tabular}
\caption{The BLEU scores of unidirectional baseline and our method with only left-to-right decoding on the three translation tasks (test sets). }
\label{tab_unidirect_compare}
\end{table}

\paragraph{Comparison with Ensemble of Two Models} We also compared our method with the ensemble of two different models, one in left-to-right decoding and the other in right-to-left decoding, with similar inference procedure as shown in Algorithm~\ref{alg}. Note that this comparison is unfair as the ensemble of two models has 2 times number of parameters as our method. We just make such comparison to demonstrate the accuracy gap between our method with a single model and the ensemble with two models. We conduct experiments on IWSLT14 De$\to$En. The BLEU score for the two unidirectional model is 32.17 (left-to-right) and 31.23 (right-to-left), and their ensemble is 33.43 while the BLEU score of our method is 33.15. There is just a minor BLEU score gap between our method and the ensemble result, considering the double model parameter size and extra training cost for the ensemble models.  Furthermore, we compare the ensemble model described above with the ensemble of two models based on our method. The BLEU score of our method with the ensemble is 33.96, still with 0.53 point gain compared with the ensemble of the baseline model.

%as follows: We first get $K$ candidates from each of the two unidirectional models (with different decoding directions) and totally we have $2K$ candidates. We rescore the candidates with both models and average the two scores on each sentence as the final reranking score. 

More analysis results of our model (varying complexity of  decoder, varying beam size in inference,  and case analysis) can be found in the supplementary materials (Section 3 and 4).

\section{Conclusions and Future Work}
In this paper, we have proposed an efficient bidirectional NMT that enables both left-to-right and right-to-left decoding in a single decoder. Experiment results demonstrate that our method can achieve significant improvements over conventional unidirectional models and be more efficient than conventional separate decoders.

For the future work, we will test our method on other sequence generation tasks, such as text summarization and image captioning. We will also investigate whether there exist some better ways to support bidirectional sequence generation in a single model, or even generate the sequence in parallel~\citep{guo2019non}.  


\bibliography{acl2019}
\bibliographystyle{acl_natbib}

\appendix
\section{Datasets Description for Neural Machine Translation}

\paragraph{IWSLT14 German-English}
We preprocess the IWSLT14 German-English (briefly, De$\to$En) dataset \citep{Cettolo2014Report} using sub-word types based on byte-pair-encoding (BPE) \citep{DBLP:conf/acl/SennrichHB16a}\footnote{https://github.com/rsennrich/subword-nmt}, resulting in a shared vocabulary of about 31K tokens. After preprocessing, the dataset contains 160K bilingual training sentences and 7K validation sentences. Following the common practice~\citep{DBLP:journals/corr/RanzatoCAZ15,bahdanau2016actor,huang2017neural},  the test set is the concatenation of dev2010, tst2010, tst2011 and tst2012, and all the tokens in the training/validation/test sentences are lower-cased. 

\paragraph{WMT14 English-German}
The training set of WMT14 English-German (briefly, En$\to$De) task consists of about 4.5 million bilingual sentence pairs\footnote{https://nlp.stanford.edu/projects/nmt/}. We use the concatenation of newstest2012 and newstest2013 as the validation set and newstest2014 as the test set. Sentences are encoded using BPE, which has a shared vocabulary of about 33000 tokens. 

\paragraph{WMT17 Chinese-English}
The training data for WMT17 Chinese-English (briefly, Zh$\to$En) task\footnote{http://www.statmt.org/wmt17/translation-task.html} consists
of 24 million bilingual sentence pairs, including CWMT Corpus 2017 and UN Parallel Corpus V1.0. We use BPE segmentation to process the source and target data, with 40K BPE tokens for the source vocabulary and 37K BPE tokens for the target vocabulary. We use the official newsdev2017 as the validation set and newstest2017 as the test set. 

\section{Experimental Setup}
\paragraph{Vocabulary} All the sentences are encoded using sub-word types based on byte-pair-encoding (BPE) \citep{DBLP:conf/acl/SennrichHB16a}\footnote{https://github.com/rsennrich/subword-nmt}. We share the vocabulary of source and target sentence in IWSLT14 De$\to$En and WMT14 En$\to$De, which results in  31K tokens and 33K tokens respectively. For WMT17 Zh$\to$En, we separate the source and target vocabulary with 40K source and 37K target BPE tokens.

\paragraph{Model Configuration} We adopt the Transformer~\citep{DBLP:conf/nips/VaswaniSPUJGKP17} as our basic model for NMT, since it achieved state-of-the-art performance in several benchmark translation tasks. For IWSLT14 De$\to$En task, we use the model configuration \textit{transformer\_small} in \citet{DBLP:conf/nips/VaswaniSPUJGKP17} with just a 2-layer encoder and a 2-layer decoder with 256-dimensional hidden representation. For WMT14 En$\to$De and WMT17 Zh$\to$En task, we use the model configuration \textit{transformer\_big} in \citet{DBLP:conf/nips/VaswaniSPUJGKP17}, which contains a 6-layer encoder and a 6-layer decoder with 1024-dimensional hidden representation.  We evaluate the translation quality by tokenized
BLEU \citep{DBLP:conf/acl/PapineniRWZ02} with multi-bleu.pl\footnote{https://github.com/moses-smt/mosesdecoder/blob/
master/scripts/generic/multi-bleu.perl} on IWSLT14 De$\to$En and WMT14 En$\to$De tasks. We use case insensitive BLEU for IWSLT14 De$\to$En and case sensitive BLEU for WMT14 En$\to$De to be consistent with previous works. Following~\citet{DBLP:conf/wmt/WangCJYCLSWY17,DBLP:conf/wmt/TanWHCS17}, we evaluate the translation quality with sacreBLEU\footnote{https://github.com/mjpost/sacreBLEU} on WMT17 Zh$\to$En task. 

\paragraph{Training and Inference}
We use the Adam optimizer \citep{kingma2014adam} with $\beta_{1}= 0.9$, $\beta_{2} = 0.98$, $\varepsilon = 10^{-9}$ and follow the same learning rate schedule in \citet{DBLP:conf/nips/VaswaniSPUJGKP17}. We train our models for WMT14 En$\to$De and WMT17 Zh$\to$En with 8 NVIDIA Tesla M40 GPUs on one machine, and for IWSLT14 De$\to$En with one M40 GPU as it is of both small model size and data size. 

During training, each mini-batch on one GPU contains a set of sentence pairs with roughly 4096 source and 4096 target tokens, which is consistent with \citet{DBLP:conf/nips/VaswaniSPUJGKP17}. In order to train our bidirectional generation model, we first gather sentence pairs with roughly 2048 source and 2048 target tokens; for each sentence pair, we reverse the target sentence on the fly, to generate training data in the reverse direction; and add them into the same mini-batch. Doing so, for each source sentence, there are both the original (left-to-right) and the right-to-left target sentences in the same mini-batch. As aforementioned, we use different start tokens indicating generation directions so as to differentiate the original target sequence and the reverse target sequence.

During inference, we followed Algorithm 1 in the paper for inference, and chose beam size $K$ and length penalty $\alpha$ according to the validation performance on each dataset.

\section{Model Analysis}

\begin{figure}[t]
\centering
\subfigure[]{
\label{sub3}
\includegraphics[width=0.22\textwidth]{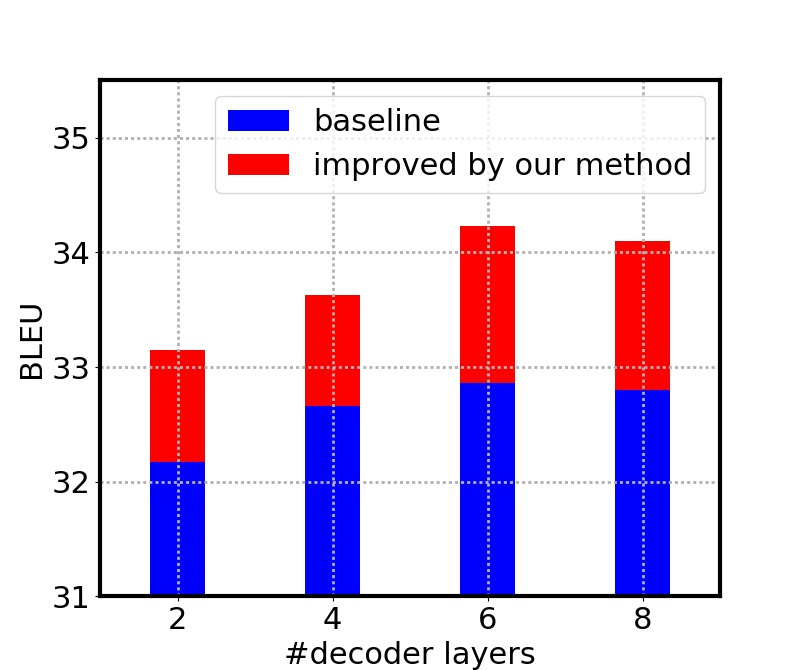}
}
\subfigure[]{
\label{sub4}
\includegraphics[width=0.22\textwidth]{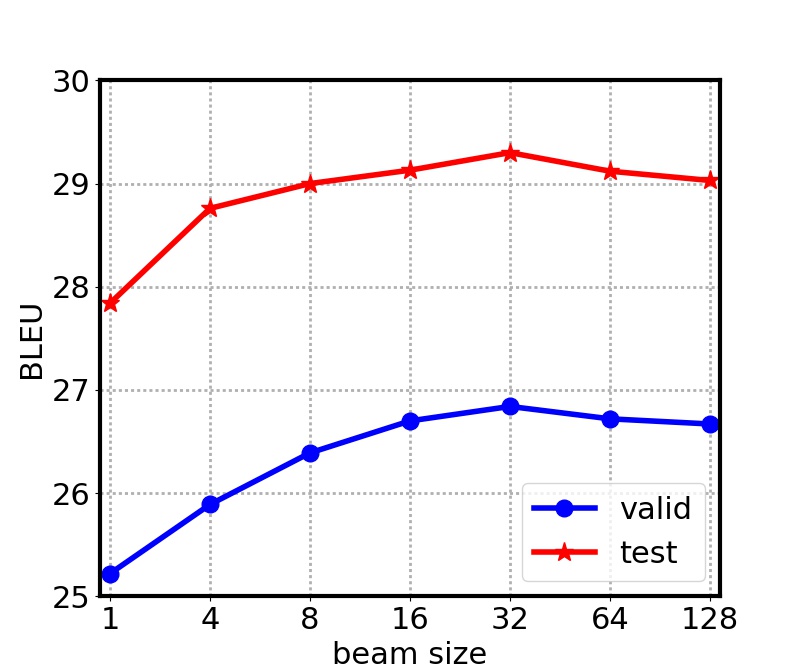}
}
\caption{(a). The test BLEU scores with respect to different number of layers in decoder on IWSLT14 De$\to$En task. (b). The validation and test BLEU scores with respect to different beam sizes on WMT14 En$\to$De task.}
\label{fig_analysis}
\end{figure}

\paragraph{Varying Complexity of Decoder}
We investigate how our approach works with respect to different complexities of the decoder. We trained a group of models with 2/4/6/8 layers in the decoder on IWSLT14 De$\to$En dataset while fixing the encoder to 2 layers. The results are shown in Figure~\ref{sub3}. We have several observations. First, our method consistently improves the BLEU scores over the baseline (the Transformer model) across different number of decoder layers. Second, our method yields more improvements as the number of layers increases from 2 to 6, achieving a BLEU score of 34.23 with 6-layer decoder. This demonstrates that better performance can be achieved by our method if increasing the complexity of the original decoder. Third, when further increasing to 8 layers, although our method can still achieve improvements over the baseline, both our method and the baseline drop. This seems to tell us that the 8-layer decoder is over complicated for this dataset.

%\begin{figure}[tbp]
%\centering 
%\includegraphics[width=0.35\textwidth]{fig/decoder_layers.jpg}
%\caption{The BLEU scores with respect to different number of layers in decoder.}
%\label{fig_dec_layer}
%\end{figure}

\paragraph{Varying Beam Size in Inference}
Beam size $K$ is a hyper parameter of our method, which decides how many candidate sequences are and could affect the performance of our method. We vary the beam size on IWSLT14 En$\to$De validation set as well as the test set to investigate its influence. The results are shown in Figure~\ref{sub4}. As can be seen, the validation BLEU increases with the beam size when the beam size is not larger than 32. Bigger beam size will result in worse BLEU score. The same pattern can be found in the test set, as shown in the blue curve in Figure~\ref{sub4}. Note that all the beam size in our paper is chosen based on the validation performance. 

%\begin{figure}[tbp]
%\centering 
%\includegraphics[width=0.35\textwidth]{fig/beam_test.jpg}
%\caption{The validation and test BLEU scores with respect to different beam sizes on the WMT14 English-German task.}
%\label{fig_beam}
%\end{figure}

\begin{table*}
\small
\centering
\begin{tabular}{l c}
\toprule
\multirow{2}{*}{\textbf{Src}} & \tl{\footnotesize es gibt keine frage, das etwas auswahl besser ist als keine, aber es folgt nicht, das mehr auswahl besser ist als}\\
& \tl{\footnotesize  etwas auswahl. }\\
\multirow{2}{*}{\textbf{Ref}} & \tl{\footnotesize there's no question that some choice is better than none, but it doesn't follow from that more choice is better} \\ 
& \tl{\footnotesize than some choice.} \\
\midrule
\textbf{L2R} & \tl{\footnotesize  there's no question that's better than a choice , but it doesn't follow the more choice \uwave{than something}.} \\

\multirow{2}{*}{\textbf{R2L}} & \tl{\footnotesize there's no question \uwave{that's a little choice better than none}, but there's question that more choice is better than} \\
& \tl{\footnotesize some choice.} \\
\midrule
\multirow{2}{*}{\textbf{Ours}}  & \tl{\footnotesize there's no question that some choice is better than none, but it's not that follows the more choice is better than} \\
& \tl{\footnotesize some choice.} \\

\midrule
\midrule
\textbf{Src} & \tl{\footnotesize hören sie bewusst zu und sagen sie, ob es ihnen schmeckt oder nicht.}\\
\textbf{Ref} & \tl{\footnotesize listen consciously and say whether you like its taste or not.} \\ 
\midrule
\textbf{L2R} & \tl{\footnotesize  listen to consciously and say whether it tastes to you or not.} \\
\textbf{R2L} & \tl{\footnotesize \uwave{listen to them} and say if you like its taste.} \\
\midrule
\textbf{Ours}  & \tl{\footnotesize listen consciously and say whether it tastes to you or not.} \\

\midrule
\midrule
\textbf{Src} & \tl{\footnotesize es ist ein unabhängiges gebiet . sie kontrollieren alle mineralvorkommen.}\\
\textbf{Ref} & \tl{\footnotesize it's an independent territory . they control all mineral resources. } \\ 
\midrule
\textbf{L2R} & \tl{\footnotesize  it's an uniquently area . they're \uwave{controlling all contributors}.} \\
\textbf{R2L} & \tl{\footnotesize it's an \uwave{inventional area} . they're controlling all the minerals.} \\
\midrule
\textbf{Ours}  & \tl{\footnotesize it's an independent area. they control all the minerals.} \\

\bottomrule
\end{tabular}
\caption{\footnotesize Translation examples on IWSLT14 De$\to$En dataset from our method (Ours) and the unidirectional results (L2R: left-to-right and R2L: right-to-left). Src/Ref represents the source/reference sentence. Text highlighted in wave line represents the improper translation.}
\label{tab:case_example_append}
\end{table*}

\section{Case Analysis}
Here we list some translation examples on IWSLT14 De$\to$En test set in Table~\ref{tab:case_example_append} below, where our method can make improvements upon the left-to-right or right-to-left unidirectional decoding.

\end{document}